\documentclass[conference]{IEEEtran}

\usepackage{cite}
\usepackage{amsmath,amssymb,amsfonts}
\usepackage{algorithmic}
\usepackage{graphicx}
\usepackage{textcomp}
\usepackage{xcolor}
\def\BibTeX{{\rm B\kern-.05em{\sc i\kern-.025em b}\kern-.08em
    T\kern-.1667em\lower.7ex\hbox{E}\kern-.125emX}}
\begin{document}

\title{Data Dimension Reduction makes ML Algorithms efficient}

\author{
% \IEEEauthorblockN{1\textsuperscript{st} Wisal Khan}
% \IEEEauthorblockA{\textit{School of Computer and Technology} \\
% \textit{Anhui University, Hefei 230039}\\
% Peoples Republic of China \\
% wisal.khan@cecos.edu.pk}
% \and
% \IEEEauthorblockN{2\textsuperscript{nd} Teerath Kumar}
% \and

\IEEEauthorblockN{1\textsuperscript{st} Wisal Khan}
\IEEEauthorblockA{\textit{School of Computer and Technology} \\
\textit{Anhui University, Hefei 230039}\\
Peoples Republic of China \\
wisal.khan@cecos.edu.pk}
\and

\IEEEauthorblockN{2\textsuperscript{nd} Muhammad Turab}
\IEEEauthorblockA{\textit{Dept: Computer Systems Engineering} \\
\textit{Mehran Uni. of Enginerring \& Technology}\\
Hyderabad, Pakistan \\
turabbajeer202@gmail.com}
\and

\IEEEauthorblockN{3\textsuperscript{rd} Waqas Ahmad}
\IEEEauthorblockA{\textit{School of Computer and Technology} \\
\textit{Anhui University, Hefei 230039}\\
Peoples Republic of China \\
waqasasad007@yahoo.com}
\and

\IEEEauthorblockN{4\textsuperscript{th} Syed Hasnat Ahmad}
\IEEEauthorblockA{\textit{School of Aeronautics} \\
\textit{Northwestern Polytechnical University}, \\ Xi'an, Shaanxi,710129, P. R of China \\
Hasnat92@hotmail.com}
\and

\IEEEauthorblockN{5\textsuperscript{th} Kelash Kumar}
\IEEEauthorblockA{\textit{Department Electrical Engineering} \\
\textit{Mehran Uni. of Enginerring \& Technology}\\
Hyderabad, Pakistan \\
kelash.nijar17@gmail.com}
\and

\IEEEauthorblockN{6\textsuperscript{th} Bin Luo}
\IEEEauthorblockA{\textit{School of Computer and Technology} \\
\textit{Anhui University, Hefei 230039}\\
Peoples Republic of China \\
luobin@ahu.edu.cn}

}

\maketitle

\begin{abstract}
Data dimension reduction (DDR) is all about mapping data from high dimensions to low dimensions, various techniques of DDR are being used for image dimension reduction like Random Projections, Principal Component Analysis (PCA), the Variance approach, LSA-Transform, the Combined and Direct approaches, and the New Random Approach. Auto-encoders (AE) are used to learn end-to-end mapping. In this paper, we demonstrate that pre-processing not only speeds up the algorithms but also improves accuracy in both supervised and unsupervised learning. In pre-processing of DDR, first PCA based DDR is used for supervised learning, then we explore AE based DDR for unsupervised learning. In PCA based DDR, we first compare supervised learning algorithms accuracy and time before and after applying PCA. Similarly, in AE based DDR, we compare unsupervised learning algorithm accuracy and time before and after AE representation learning.  Supervised learning algorithms including support-vector machines (SVM), Decision Tree with GINI index, Decision Tree with entropy and Stochastic Gradient Descent classifier (SGDC) and unsupervised learning algorithm including K-means clustering, are  used for classification purpose. We used two datasets MNIST and FashionMNIST Our experiment shows that there is massive improvement in accuracy and time reduction after pre-processing in both supervised and unsupervised learning.
\end{abstract}


\section{Introduction}
Machine learning (ML) algorithms are successful in range of domain tasks such as image classification~\cite{kumar2021binary,kumar2021class,kumarforged,khan2022introducing}, audio classification ~\cite{chandio2021audd,park2020search,kumar2020intra,turab2022investigating}  and data dimension reduction \cite{lee1993analyzing,li2011locality} and many more ~\cite{kumarstride}. Among them,  
DDR reduces \cite{lee1993analyzing,li2011locality} the high dimension representation into low level representation  for data processing step.  In the current era, image classification is the core task and improving accuracy and time are the key factors in image classification. The higher image dimension, the more time machine learning algorithms require to process and at the same time, it is very difficult for the algorithms to classify due to higher dimension. To cater these issues, we found that data dimension reduction (DDR) can solve these issues. For the purposes, we use AE and PCA for data dimension reduction and show their effect. 

% and mostly machine learning algorithms are heavily reliant on labeled data  is their core task. But manually labeling the data takes a lot of time. Here comes clustering to label the data but we know it is a waste of time and useless to apply clustering directly on image, savior encoder save time and improve accuracy.
\subsection{Auto-Encoders (AE)}
\begin{figure}[htbp]\centerline{\includegraphics[width=0.5\textwidth]{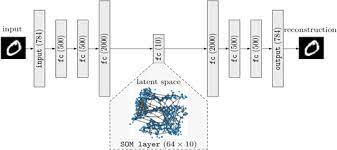}}
    \caption{Auto-encoder (Encoder-Decoder) architecture. Image taken from ~\cite{teimoorinia2021astronomical}}
    \label{fig1}
    \end{figure}

Convolutional Neural Network (CNN) is one of the established deep learning techniques and brought certain advantages in feature extraction of high level semantics in the field of image detection, segmentation and classification. Considering image feature extraction many researchers have been proposed small optimized convolutional neural networks\cite{krizhevsky2012imagenet,simonyan2014very,he2016deep}. The purpose of the Auto-Encoders (AE) is to copy their input to their output along with many constraints. The AEs\cite{autoencoders} are neural networks. The main task of AE is the input compression to gain the latent space representation. And then this representation is used to build the desired output. The two main components of this kind of network are: Encoder and Decoder. The job of the AEs is not only to map the input to the output but also to learn latent space representation. The two main practical applications of AEs are data denoising and dimensionality reduction for data visualization. AEs are categorized into 4 different types to achieve various properties when input is passed via encoder and decoder to construct a new presentation. The types are Vanilla, Multilayer, Convolutional, and Regularized AEs. The representation of image input is described in figure~\ref{fig1}.

% \begin{figure}[htbp]
% \centerline{\includegraphics[width=0.5\textwidth]{images/fig1.png}}
% \caption{Encoding and decoding structure. Figure Taken from ~\cite{Francois16}}
% \label{fig}
% \end{figure}

% The researchers gave much attention to the image retrieval from last decade and the active and hot area of research. Google Image Search and Pinterest Visual Pin Search are the most famous systems. They built an image retrieval system via AEs and followed unsupervised learning approach. Unsupervised learning means images labels are not considered. The image retrieval via mentioned procedure is known as Content-based Image Retrieval (CBIR)\cite{nathanretrievalsystem}.
\subsection{Principal Component Analysis (PCA)}
Principal Component Analysis (PCA) is another technique to reduce image dimensions\cite{moore1981principal,jolliffe2002principal}. PCA is based on mathematical functions. Single value decomposition method is used by the PCA\cite{tzeng2013split} to extract the essential features of a linear system. In many fields PCA is being widely applied such as digital signal processing, image recognition as well as classification problems to discard noisy data\cite{petrolis2013multi,shi2009application,tzimiropoulos2011principal,li2016accurate,kamencay2016accurate}. PCA technique is applied to medical data to compress digital images according to the study of\cite{santo2012principal}.

Labeling data is tedious, time-consuming and expensive task. In order to limit these problem, we explore DDR for unsupervised learning that would cluster similar type of objects. But we know, applying clustering directly on images is useless and waste of time. Our main contribution is, we pre-processed data using state-of-the-art (SOTA) encoder-decoder models to get better representation for making accurate and efficient prediction. We also compare with other SOTA techniques such as PCA. So our main focus on pre-processing of data and how it helps in speeding up algorithm and improving accuracy for clustering similar type of objects. And further it can be used for auto labeling the data. The following is the contribution of our proposed work:
\begin{itemize}
\item We demonstrate how pre-processing helps in improving speed and accuracy in both supervised and unsupervised learning.
\item We show the effect of DDR for supervised and unsupervised learning accuracy and time duration improvement.
\item We put our analysis on why accuracy and speed improved and clustering for auto labelling the data.
\end{itemize}
The rest of the paper is organized as follows.Section ~\ref{r_Work} describes related work, section \ref{methdlogy} explain methodology, section \ref{experiments} defines the experimental setup, section \ref{results} discusses results evaluations and finally section \ref{conclusion} concludes the whole work.

\section{Related Work}\label{r_Work}
Retrieving the image, the non-linearity can be learned by multiple layers of CNN to extract and characterize features. The article \cite{krizhevsky2012imagenet} investigated that by observing the advancement in the CNN regarding image segmentation, classification and detection we can use for image retrieval the deep learning approach. They\cite{krizhevsky2012imagenet} achieved good result on ImageNet. For image retrieval,  they learn similarity using different networks across different images in dataset, query image features are calculated and the final results are visualized~\cite{giveki2022new}. The study\cite{xia2014supervised} proposed the CNNH (Convolutional Neural Network Hashing) for learning more robust features. CNNH first decomposed the similarity matrix by using the similarity between the training images. The training image binary coding is then obtained via similarity matrix. The obtained binary coding is then used by the CNN. Considering image retrieval paper\cite{wan2014deep} performed characteristics learning techniques on images by using CNN. DSRH (Deep Semantic Ranking Hashing) is investigated by article\cite{zhao2015deep}. According to DSRH, the image retrieval job is to transform into solving the problem of image relevancy ranking. Image learning feature representation is achieved via deep convolutional neural network (CNN) in DSRH. The obtained learning features of an image are then mapped to a hash code. The study\cite{lai2015simultaneous} investigated a technique for simultaneous feature learning and hash coding via deep convolutional neural network. According to their proposed method\cite{lai2015simultaneous} the image features are divided into individual encoding module into many chunks. In the hash code each chunk is responsible for learning one bit. In the paper\cite{liu2016deep} the image pair supervision information has been utilized. The deep neural network has been constrained via a proposed regular term. Therefore, the neural network output is close to the binary code.   In computer vision, according to the literature, the CBIR is one of the hot areas of the last decade. The studies\cite{krizhevsky2012imagenet,simonyan2014very,he2016deep} describes that for feature extraction from image search based on deep learning used classical classification models. The performance has been improving of these classification networks but too large models. Besides the large models the extracted feature dimension is also too high. Therefor for image search applications it is still insufficient. 
In CNN the latter two layers are good for image search feature retrieval as investigated by\cite{wan2014deep}.
According to the research the authors of~\cite{krizhevsky2011using} are the founders of depth Auto-Encoder for image retrieval. They~\cite{krizhevsky2011using} searched in CIFAR10 dataset and gained some good outcomes by using Boltzmann machine to encode and decode. The image features are compressed to 254 bits. Extracting features via neural networks the article\cite{babenko2014neural} and several other researchers used PCAs. IBM-supervised KL divergence is introduced by the author of\cite{liuhashing} and is used for extracting the features. Approximate nearest neighbor approach is used instead of linear search to speed up the image retrieval\cite{lin2015deep}.
In the regarding of computer vision similarity, many proposed methods are investigated by the researchers such as approaches based on hashing\cite{gionis1999similarity,weiss2008spectral,liu2012supervised,norouzi2011minimal,zeiler2013stochastic}.
Hamming distance is used by the hashed-based approaches to calculate the similarity. The authors of\cite{qin2018image} investigated a hybrid model for CBIR. By considering extracting the corresponding image features, the authors\cite{qin2018image} used convolutional neural network and depth Auto-Encoder. 
% \begin{figure}[htbp]
% \centerline{\includegraphics[width=0.5\textwidth]{images/fig2.png}}
% \caption{Top 10 image retrieval results for the CIFAR10 dataset. The leftmost one is the query input image, and the rightmost 10 are the results. Image taken from ~\cite{krizhevsky2009learning}}
% \label{fig}
% \end{figure}
According to the literature the following compression coding techniques and well established used in JPEG\cite{wu2014genetic,hussain2015hybrid,prabhu20133,quijas2014removing} standard are Fast Fourier Transform (FFT), Discrete Cosine Transform (DCT), Artificial Neural Network, and Discrete Wavelet Transform (DWT). The PCA technique is explained by the study\cite{santo2012principal,bao2012inductive} in detail. The article\cite{qin2018image} explains the digital image reduction process via PCA and used following equations.

\begin{equation}
F(x, y)=\left[\begin{array}{ccc}
f(0,0) & \cdots & f(0, m-1) \\
\vdots & \ddots & \vdots \\
f(n-1,0) & \cdots & f(n-1, m-1)
\end{array}\right]
\label{eq1}
\end{equation}

In equation ~\ref{eq1}, F(x,y) denoting the required colors while x, y represents the pixels coordinates of the image. Equation~\ref{eq2} is about image normalization.

\begin{equation}
\resizebox{1.0\hsize}{!}{
$F_{\text {normalized }}(x, y)=\left[\begin{array}{ccc}
f(0,0) & \cdots & f(0, m-1) \\
\vdots & \ddots & \vdots \\
f(n-1,0) & \cdots & f(n-1, m-1)
\end{array}\right]-[\bar{f}(0,0) \quad \ldots \quad \bar{f}(0, m-1)]$
}
\label{eq2}
\end{equation}

Equation ~\ref{eq3} is representing the Covariance Matrix of image data.

\begin{equation}
\operatorname{Cov}(X, Y)=\frac{F_{\text {normalized }}(x, y) * F_{\text {normalized }}(x, y)^{T}}{m-1}
\label{eq3}
\end{equation}

The Covariance Matrix Eigenvectors and Eigenvalues are given in equation \ref{eq4} while equation \ref{eq5} representing the image transformation data into New Basis.

\begin{equation}
\resizebox{0.7\hsize}{!}{
$AA^{T}=\operatorname{Cov}(X, Y)=U D^{2} U^{T}$
}
\label{eq4}
\end{equation}

\begin{equation}
\resizebox{0.8\hsize}{!}{
$F_{\text {transformed }}(x, y)=U^{T} F_{\text {normalized }}(x, y)$
}
\label{eq5}
\end{equation}
In equation~\ref{eq4} and equation~\ref{eq5}, $U$, $U^T$ and $D^2$ are matrices. Column vectors of $U$ represent eigen vector while diagonal entr(ies) of $D^2$ represent eigen values. 
% Proposed method
\section{Methodology}\label{methdlogy}
We have divided method into two sections one with supervised learning and other with unsupervised learning. We will show comparison results within the experiments section.

\subsection{Supervised}\label{1}
In this part, we used first PCA algorithm to reduce the dimension of data (images) and then applied some machine learning algorithms\cite{dimensionreduction}, and showed how pre-processing speedup algorithm and with comparable accuracy as shown in experiments section. 
For this supervised learning to learn faster, we first projected the data into low dimension using PCA.

% % algo
% \begin{figure}[htbp]
% \centerline{\includegraphics[width=0.5\textwidth]{images/fig4.jpg}}
% \end{figure}
% % end algo

After applying PCA, we used below machine learning algorithm to check accuracy and time before and after PCA applying.

\begin{itemize}
    \item SVM (with and without PCA) 
    \item Decision Tree with GINI index (with and without PCA)
    \item Decision Tree with entropy (with and without PCA)  
    \item Stochastic Gradient Descent classifier (with and without PCA)
    \item Gaussian Naive Bayes (with and without PCA)
\end{itemize}

\subsubsection{\textbf{SVM}}
Support Vector Machine (SVM) is a classifier that is used for binary or multi classification problem\cite{svmsavan}. Technically, provided labels with data, it learns the parameter and calculate the optimal hyperplane.  In SVM, kernel plays very important role.  For linear kernel the equation for prediction for a new input as follows: 

\begin{equation}
\mathrm{F}(\mathrm{x})=\mathrm{w}^{*} \mathrm{x}+\mathrm{b}
\label{eq6}
\end{equation}

Equation~\ref{eq6} used in SVM, where we learn w and b from training data, once these are learnt they can be used for testing.
The polynomial kernel can be written as
\begin{equation}
K(x, x i)=1+\operatorname{sum}(x * x i)^{d}
\end{equation}
and exponential as
\begin{equation}
K(x, xi) = exp (-gamma * sum ((x-xi^{2})) 
\end{equation}.
\subsubsection{\textbf{Decision Tree with GINI index}}
Decision tree is classification algorithm used for classification and regression both. They are very popular because trees often mimic the human level thinking so easy to interpret the data.  In Classification with using the Classification and Regression Trees (CART) algorithm, mostly GINI index is used. 
Gini impurity is a measurement of how often a randomly chosen element from the set would be incorrectly labeled if it was randomly labeled according to the distribution of labels in the subset\cite{ginindex}.

\subsubsection{\textbf{Decision Tree with Entropy}}
Decision tree is built as top-down from root node and composed of partitioning the data into subsets that has further many instances with same value that we call as homogeneous. Entropy means disorder, and it measures the homogeneity of sample. If the sample is completely homogeneous, then entropy is zero and if the sample is equally divided then it has entropy of one\cite{decisiontrees}.

% We have found two types of entropy in decision as follow:
% \begin{itemize}
%     \item Entropy using the frequency table of one attribute: Below equation with example.
    
%     \begin{figure}[htbp]\centerline{\includegraphics[width=0.5\textwidth]{images/fig5.png}}
%     \caption{Entropy using frequency table. Image is taken from ~\cite{SUPRIYA18}}
%     \label{fig}
%     \end{figure}

%     \item Entropy using the frequency table of two attributes: Below equation with example
%     \begin{figure}[htbp]\centerline{\includegraphics[width=0.5\textwidth]{images/fig6.jpg}}
%     \caption{Entropy using the frequency table of two attributes. Image is taken from ~\cite{SUPRIYA18}}
%     \label{fig}
%     \end{figure}
% \end{itemize}

\subsubsection{\textbf{Stochastic Gradient Descent Classifier}}
The stochastic gradient method is a gradient decent method used to classify the data point and highly depends on rate of the convergence. It differs from traditional gradient in a way; its elements are interpreted separately. It approximate the gradient while using only one data point and results in saving a lot of time, and really more helpful while working with huge dataset. Procedure is described below.

\begin{itemize}
    \item Shuffle dataset randomly
    \item Cycle on all elements of the sample
    \item Cycle on all weights
    \item Adjust the current weight in accordance with the private derivative of the cost function.
\end{itemize}

It is really critical to get to know that in contrast to traditional method of gradient descent, this algorithm at every step may not endeavor to minimize the cost function, but as a result of specified steps number, the general direction will tend to this minimum\cite{sgdhakob}.

\subsubsection{\textbf{Gaussian Naive Bayes}}
First let’s know Naïve Bayes, it is classifier based on probability, it calculates the probability of each example or sample be part of certain label or category, based on prior knowledge\cite{saranaive}. 

% \begin{figure}[htbp]\centerline{\includegraphics[width=0.5\textwidth]{images/fig7.png}}
%     \caption{Naïve Bayes theorem. Image is taken from \cite{saranaive}}
%     \label{fig}
%     \end{figure}
% In Gaussian Naïve Bayes, distribution of probability is assumes as Gaussian.

\subsection{\textbf{Unsupervised}}
In unsupervised learning\cite{Chengweiclustering}, we used first reduced image dimension by training encoder decoder and then apply k-means clustering to know improved accuracy. Below is architecture of encoder decoder architecture.

\begin{figure}[htbp]\centerline{\includegraphics[width=0.5\textwidth]{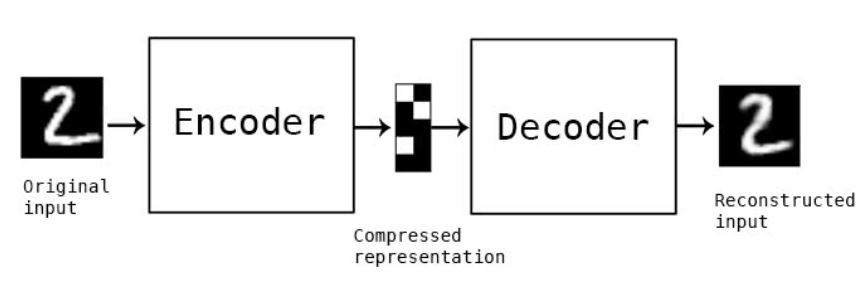}}
    \caption{Encode decoder architecture. Image is taken from ~\cite{Francois16}}
    \label{fig3}
    \end{figure}

From above figure ~\ref{fig3}, we first pass train image twice, one time as an image to be processed and secondly as a label to be learnt. We want to ensure that encoder and decoder learnt compressed representation properly or not. Once it is trained, then we compress all train images and then apply k-means clustering. We showed accuracy and duration of clustering after and before image compression. Figure~\ref{fig1} and figure~\ref{fig3} represent white box  and black box architecture, respectively. .

\section{Experiment}\label{experiments}
For our experiment, we used two datasets, MNIST and FASHION-MNIST. For each dataset, we applied both supervised and unsupervised learning. First we will explain the supervised learning, in which we first reduced dimension (used 25 components) of image using PCA then applied five supervised learning algorithms. In this experiment, we will show accuracy and time before and after PCA.  For unsupervised learning, first we trained the encoder decoder to learn image in compressed form, then applied k-means clustering. In this experiment, we will show accuracy and time before and after compressed form images. At the end we will show, how much unsupervised learning is close to supervised learning in the term of accuracy.

\begin{table*}[htp!]\small 
\centering
\label{architecture}
\begin{tabular}{|l|l|l|l|l|l|}
\hline  & \multicolumn{2}{c|}{\textbf{Accuracy}} &  \multicolumn{2}{c|}{\textbf{Time}} \\
 
\hline \textbf{Algorithms} & \textbf{Before PCA} & \textbf{After PCA} & \textbf{Before PCA} & \textbf{After PCA} \\
\hline Decision Tree GINI index & 85.44\% & 81.59\% & 18.649s & 2.152s  \\
\hline Decision Tree Entropy & 88.05\% & 84.1\% & 27.447s & 8.664s  \\
\hline Stochastic Gradient Descent classifier & 85.44\% & 85.22\% & 5.491s & 0.321s \\
\hline Gaussian Naive Bayes & 55.58\% & 86.38\% & 0.585s & 0.041s  \\
\hline SVM & --- & 96.77\% & --- & 163.970s \\
\hline
\end{tabular}
\caption{Supervised Learning for MNIST dataset}
\label{table1}
\end{table*}

\begin{table*}[htp!]\small 
\centering
\label{architecture}
\begin{tabular}{|l|l|l|l|l|l|}

\hline  & \multicolumn{2}{c|}{\textbf{Accuracy}} &  \multicolumn{2}{c|}{\textbf{Time}}\\
\hline \textbf{Algorithm} & \textbf{Before PCA} & \textbf{After PCA} & \textbf{Before PCA} & \textbf{After PCA} \\
\hline Decision Tree GINI index & 80.44\% & 76.06\% & 32.101s & 3.233s  \\
\hline Decision Tree Entropy & 81.58\% & 78.39\% & 39.973s & 11.347s \\
\hline Stochastic Gradient Descent classifier & 82.28\% & 78.06\% & 5.385s & 0.380s \\
\hline Gaussian Naive Bayes & 58.56\% & 75.07\% & 0.842s & 0.115s \\
\hline SVM & -- & 84.79\% & --- & 162.740s\\
\hline

\end{tabular}
\caption{Supervised Learning for FASHION-MNIST dataset}
\label{table2}
\end{table*}
\section{Results Evaluations}\label{results}
\subsection{\textbf{Supervised learning experiment result}} In this experiment,  we used MNIST and FASHION-MNIST datasets and their results are shown in tables \ref{table1} \& \ref{table2} respectively. Note: --- means, it was unable to find decision boundary and it took too much time to train.

\begin{figure}[htbp]\centerline{\includegraphics[width=0.5\textwidth]{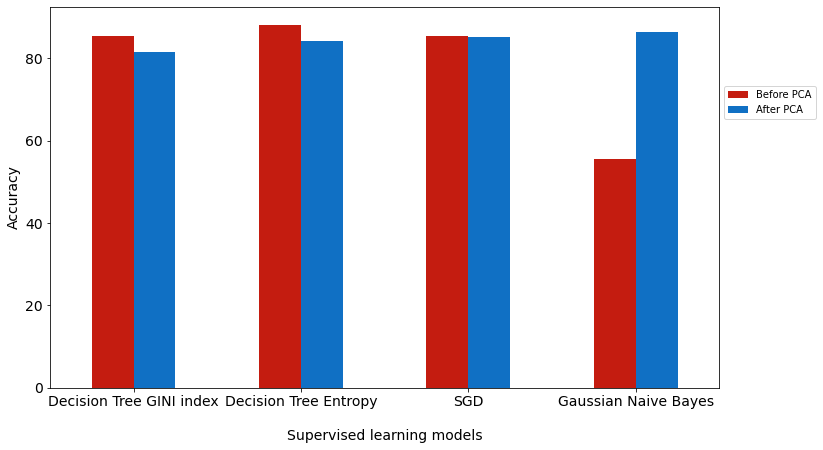}}
    \caption{Supervised learning accuracy results for MNIST dataset}
    \label{fig33}
    \end{figure}

\begin{figure}[htbp]\centerline{\includegraphics[width=0.5\textwidth]{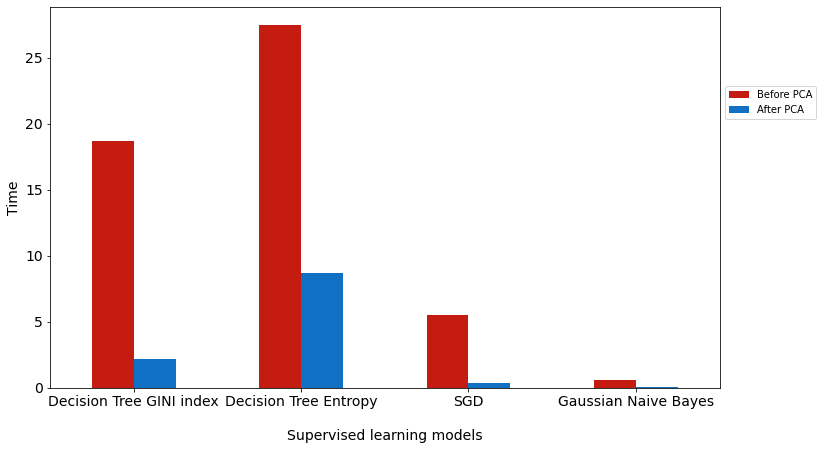}}
    \caption{Supervised learning time results for MNIST dataset}
    \label{fig4}
    \end{figure}

\begin{figure}[htbp]\centerline{\includegraphics[width=0.5\textwidth]{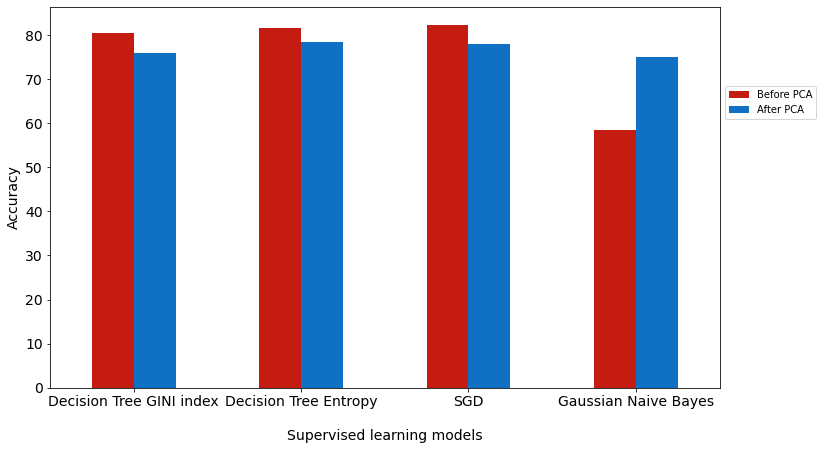}}
    \label{fig5}
    \end{figure}

\begin{figure}[htbp]\centerline{\includegraphics[width=0.5\textwidth]{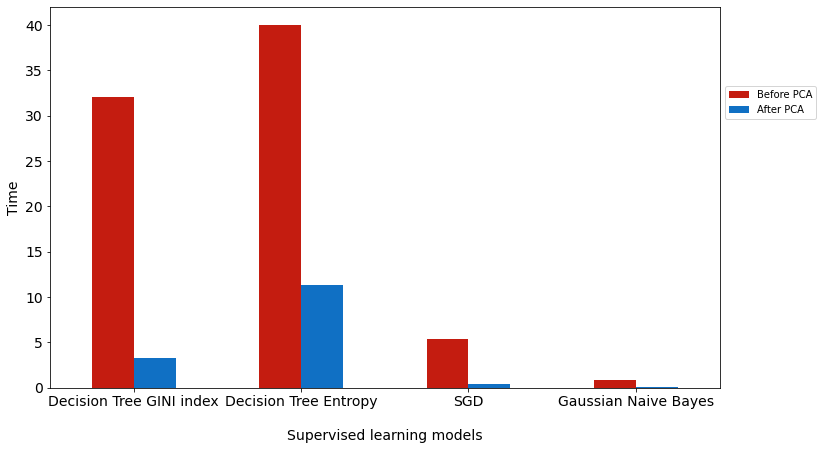}}
    \caption{Supervised learning time results for Fashion-MNIST dataset}
    \label{fig6}
    \end{figure}

As you can see that how much time is reduced in some cases, it is reduced to nine times plus relatively. Regarding, accuracy, PCA reduction outperforms before PCA cases.  For better understanding, we visualised the result before and after dimension reduction, for MNIST and FashionMNIST datasets for time and accuracy as shown in figures \ref{fig33}, \ref{fig4} \ref{fig5} and \ref{fig6}. Figure~\ref{fig6} shows major time difference  before and after PCA. It can be easy for PCA to learn better representation and additionaly, it can also be dependent on model as different model is showing different gap before and after PCA. From this experiment, we conclude that pre-processing not help in improving time but also accuracy. 

\subsection{\textbf{Unsupervised learning experiment}}
In this experiment, we first apply k-means clustering directly on images and we get lower accuracy. To improve the accuracy, we trained the encoder to encode an image in low dimension and then apply k-means clustering. Here we refer Normalize mutual information score as accuracy. Encoder is not helpful in accuracy improving but it also time efficient as shown in below tables ~\ref{table3} \& ~\ref{table4} for MNIST and FASHION-MNIST dataset respectively.

\begin{table*}[htp]\small 
\centering
\label{architecture}
\begin{tabular}{|l|l|l|l|l|l|}
\hline  & \multicolumn{2}{c|}{\textbf{Accuracy}} &  \multicolumn{2}{c|}{\textbf{Time}} \\
\hline \textbf{Algorithm} & \textbf{Before Encoder} & \textbf{After Encoder} &  \textbf{Before Encoder} & \textbf{After Encoder} \\
\hline K-means & 49.81\% & 75.37\%  & 205.714s & 7.5889  \\
\hline
\end{tabular}
\caption{Unsupervised Learning for MNIST dataset}
\label{table3}
\end{table*}

\begin{table*}[htp]\small 
\centering
\label{architecture_fmnist}
\begin{tabular}{|l|l|l|l|l|l|}
\hline  & \multicolumn{2}{c|}{\textbf{Accuracy}} &  \multicolumn{2}{c|}{\textbf{Time}} \\
\hline \textbf{Algorithm} & \textbf{Before Encoder} & \textbf{After Encoder}  & \textbf{Before Encoder} & \textbf{After Encoder}  
\\
\hline K-means & 51.4\% & 58.56\% & 224.386s & 8.597s  \\
\hline
\end{tabular}
\caption{Unsupervised Learning for FASHION-MNIST dataset}
\label{table4}
\end{table*}

From this experiment, we have shown that pre-processing not helps in supervised learning, but also in unsupervised learning. As above both table showed, not only time reduced but also accuracy improved a lot. It is improved because of low dimension of data representation, that low dimension data become easy for k-means algorithm to learn and make cluster easily. For cluster evaluation we used normalized mutual information score.

% \begin{figure}[htbp]\centerline{\includegraphics[width=0.5\textwidth]{images/fig10.jpg}}
%     \caption{Mutual information score}
%     \label{fig}
%     \end{figure}

\subsection{Supervised and unsupervised learning comparison}
One common thing is preprocessing helped in both learning a lot. We will discuss accuracies and durations of both learning after preprocessing. In supervised learning, accuracy is improved a little bit and in some cases it is reduced a little bit too. But in clustering, it is improved in both cases a lot. Because for clustering it becomes easy to learn data in low dimension, so clustering accuracy is much closer to the supervised learning algorithms. Regarding duration, it has improved a lot in both learning after preprocessing.

\textbf{Analysis:Why dimension helpful?} This is because of pre-processing, DDR reduced the dimensions and it becomes easy for algorithms to classifier and at the same time, it takes a very less time, that is why algorithm learns faster and more accurate.

\section{Conclusion}\label{conclusion}
In this paper, we demonstrated that the data dimension reduction helps to improve accuracy and time of ML algorithms. To do so, we explored two different technique of data dimension reduction for supervised and unsupervised learning using two different datasets namely, MNIST and Fashion-MNIST. DDR has shown massive improvement in term of time and accuracy. Furthermore, we discussed result behind efficiency of these algorithm due to DDR. Our future work is to explore these techniques for data other than gray. 

\bibliographystyle{plain}
\bibliography{references.bib} 
\end{document}